\documentclass[a4paper,conference]{IEEEtran}
\ifCLASSINFOpdf
   \pdfoutput=1
\else
\fi
\usepackage{amsmath}

\usepackage{afterpage}

 \usepackage[caption=false,font=normalsize,labelfont=sf,textfont=sf]{subfig}
\usepackage{stfloats}
\usepackage{url}

\usepackage{tikz}
\usetikzlibrary{positioning,shapes,shadows,arrows,backgrounds,fit,calc}

\definecolor{amethyst}{rgb}{0.6, 0.4, 0.8} 
\definecolor{amber}{rgb}{1.0, 0.49, 0.0} 
\definecolor{alizarin}{rgb}{0.82, 0.1, 0.26} 
\definecolor{applegreen}{rgb}{0.55, 0.71, 0.0} 
\definecolor{bananayellow}{rgb}{1.0, 0.88, 0.21} 
\definecolor{apricot}{rgb}{0.98, 0.81, 0.69} 

\usepackage{color}

\begin{document}
%
\title{RescueNet: Joint Building Segmentation and Damage Assessment from Satellite Imagery }

\author{\IEEEauthorblockN{Rohit Gupta and Mubarak Shah}
\IEEEauthorblockA{Center for Research in Computer Vision\\
University of Central Florida\\
Orlando, Florida 32816\\
Email: rohitg@knights.ucf.edu, shah@crcv.ucf.edu}
}


%


\maketitle

\begin{abstract}
Accurate and fine-grained information about the extent of damage to buildings is essential for directing Humanitarian Aid and Disaster Response (HADR) operations in the immediate aftermath of any natural calamity. In recent years, satellite and UAV (drone) imagery has been used for this purpose, sometimes aided by computer vision algorithms. Existing Computer Vision approaches for building damage assessment typically rely on a two stage approach, consisting of building detection using an object detection model, followed by damage assessment through classification of the detected building tiles. These multi-stage methods are not end-to-end trainable, and suffer from poor overall results. We propose RescueNet, a unified model that can simultaneously segment buildings and assess the damage levels to individual buildings and can be trained end-to-end. In order to to model the composite nature of this problem, we propose a novel localization aware loss function, which consists of a Binary Cross Entropy loss for building segmentation, and a foreground only selective Categorical Cross-Entropy loss for damage classification, and show significant improvement over the widely used Cross-Entropy loss. RescueNet is tested on the large scale and diverse xBD dataset and achieves significantly better building segmentation and damage classification performance than previous methods and achieves generalization across varied geographical regions and disaster types.

\end{abstract}


%
\IEEEpeerreviewmaketitle

\section{Introduction}

\begin{figure}[!h]
\centering
\subfloat[Pre-Disaster]{\includegraphics[width=1.6in]{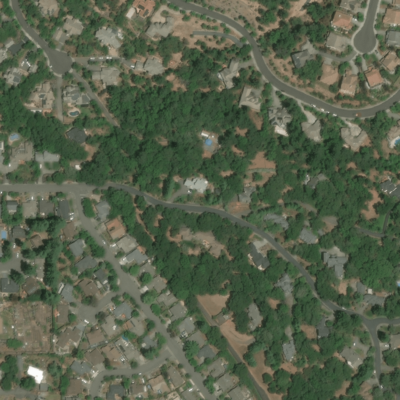} \label{fig:preimg}}
\hfill
\subfloat[Post-Disaster]{\includegraphics[width=1.6in]{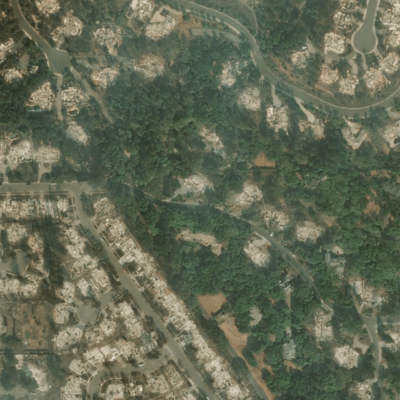} \label{fig:postimg}} \\
\subfloat[Ground Truth]{\includegraphics[width=1.6in]{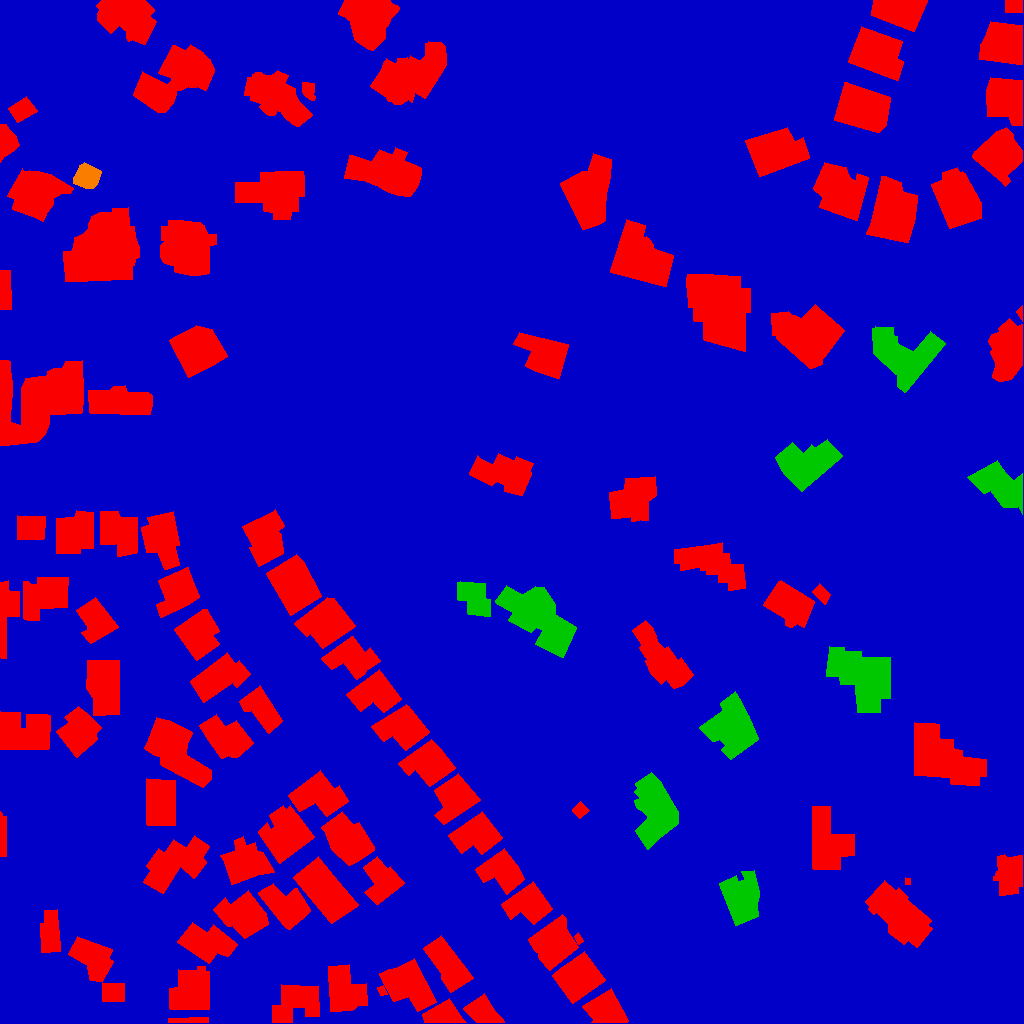} \label{fig:gt}}
\hfill
\subfloat[Prediction (Ours)]{\includegraphics[width=1.6in]{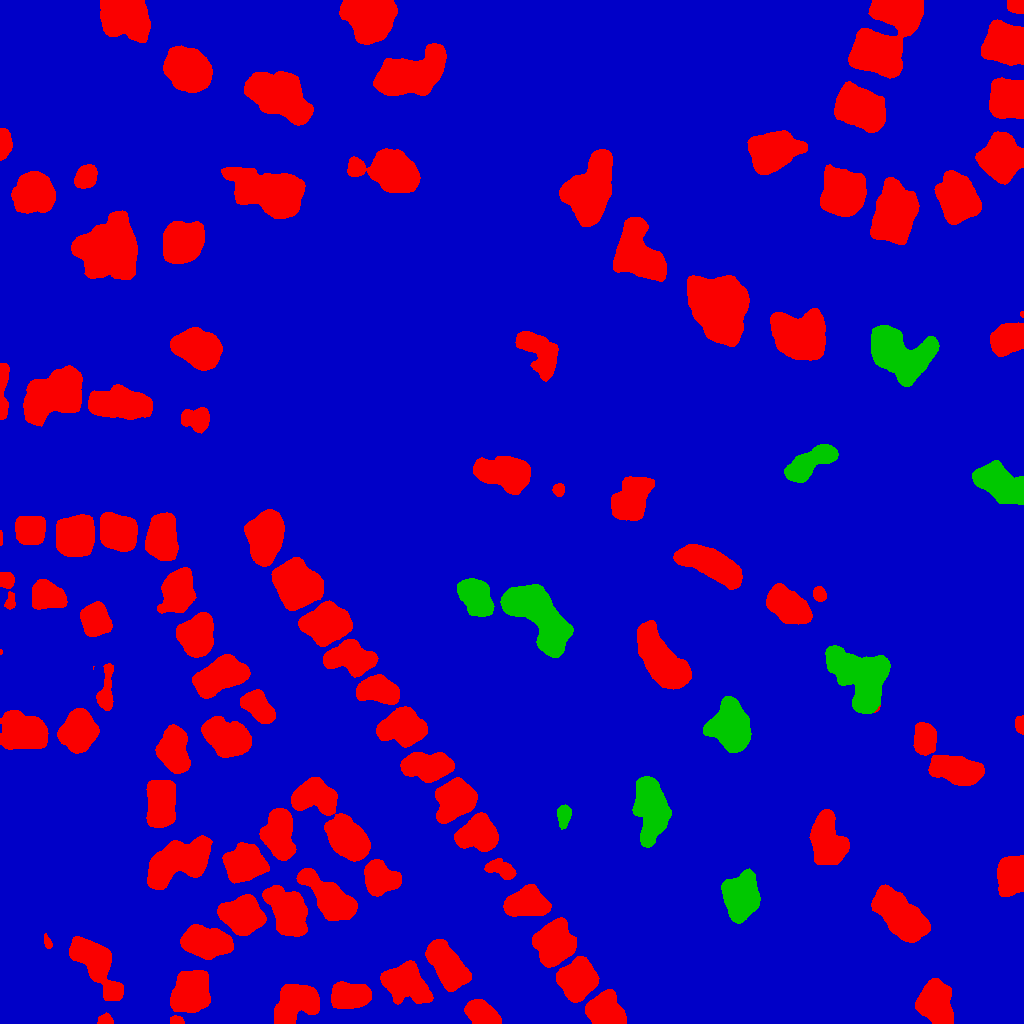} \label{fig:pred}} \\
\caption{Samples from the xBD dataset \cite{gupta2019xbd} for building damage assessment. (a) and (b) respectively show Pre and Post Disaster images. (c) and  shows ground truth and prediction from proposed method. Red labels represent completely damaged buildings, and Green labels represent undamaged buildings. Intermediate damage levels are represented by pink (major damage) and orange (minor damage).}
\label{fig:sample}
\end{figure}

In the wake of natural disasters, resources available to first responders are scarce, and efficient planning and allocation of aid and rescue efforts can help save thousands of lives. Traditionally, response planning has been based on reports and estimates based on ground based assessments. Ground based assessments are risky and potentially impossible to obtain, hence, more recently aerial and satellite imagery has been used for these assessments \cite{disasters2018}. While analysis of satellite and aerial imagery by experts is useful for rapid response operations, it still results in time lags that could otherwise be spent on rescue operations, as even large teams can take weeks to completely map out disaster affected areas \cite{foulser2012use}. Automated methods for analyzing aerial and satellite imagery have been developed, including those relying on handcrafted rules for identifying damaged buildings from LiDAR point clouds \cite{axel2017building}, segmenting the perimeter of forest fires using deep learning \cite{doshi2019firenet}, detecting flooded regions \cite{mateo2019flood}, detecting collapsed, and damaged buildings using Convolutional Neural Networks \cite{cooner2016detection} \cite{nex2019structural} and detecting damaged buildings using object detectors \cite{li2019building}. 

In this paper, we focus on assessing damage levels for buildings, which is relevant to all types of natural disasters, and can have a significant impact on the efficacy of search and rescue operations in their aftermath. State of the art methods for detecting damaged buildings \cite{xu2019building} \cite{gupta2019xbd} rely on a two stage pipeline, where buildings are detected in the pre-disaster imagery in the first stage, and then detected building are classified into different damage levels by comparing pre and post disaster imagery in the second stage. These multi-stage methods are not end-to-end trainable, and suffer from poor overall results. In contrast, we propose RescueNet, which is an end-to-end trainable, unified model to segment buildings and classify their damage levels in one go. We employ a pixel-level segmentation based approach, and use multi-scale, temporal (capturing before and after damage) features to detect damage in the segmented buildings. To model the hierarchical and composite nature of this problem, we also design a novel loss function which provides a significant improvement over the standard cross-entropy loss. To benchmark our method, we compare against an existing baseline on the public and large scale xBD dataset \cite{gupta2019xbd}, which contains images from a variety of disaster events such as earthquakes, flooding, hurricanes and forest fires from 19 locations across the world. This diversity allows for training and testing the model on a wide variety of damage types. RescueNet is able to achieve significantly improved performance over the baseline, especially on damage classification. Sample results are shown in Figure \ref{fig:sample} and more results are show in Figure \ref{fig:qual}.




The organization of the rest of the paper as follows. Section \ref{relatedwork} provides a brief overview of the existing literature on the problems of Building Footprint Segmentation, Building Damage Assessment and Change Detection in remote sensing using satellite imagery. Section \ref{model} presents the architecture of our model, and our training approach. Section \ref{dataset} provides a brief summary of the characteristics of the xBD dataset. Section \ref{results} contains an analysis of the results obtained by our method. Finally, our conclusions are provided in Section \ref{conclusion}.

\section{Related Work} \label{relatedwork}

\subsection{Building Footprint Segmentation}

Most prior works on training deep building footprint detection models use the DeepGlobe \cite{demir2018deepglobe} or SpaceNet \cite{van2018spacenet} datasets, while others like BingHuts \cite{marcos2018learning} and the ISPRS 2D Semantic Labelling  (Vaihingen and Potsdam) \cite{isprssemantic} dataset also exist.  Whereas, SpaceNet and DeepGlobe are more diverse and large scale, hence better suited for training and evaluating deep learning methods. State of the art methods for Building footprint segmentation on these datasets typically rely on models like U-Net \cite{ronneberger2015u}, originally developed for semantic instance segmentation problems. TernausNetv2 \cite{iglovikov2018ternausnetv2} is a modified U-Net  to operate on multi-spectral imagery, and to predict instance boundaries along with pixelwise semantic segmentation. Hamaguchi \textit{et al}. \cite{hamaguchi2018building} utilize ensembles of U-Nets, each of which is trained to detect buildings of a specific size range. A variant of LinkNet (a U-Net with residual connections in the encoder),  \cite{chaurasia2017linknet} with SE-ResNext backbone, and a composite loss function consisting of the binary cross-entropy loss, combined with the Lov\'{a}sz hinge loss and mean squared watershed energy loss is used in Golovanov \textit{et al}. \cite{golovanov2018building}.

\subsection{Change Detection}

There has been an extensive amount of research on change detection by remote sensing researchers prior to the advent of deep learning. Typically, these methods applied pixel difference based models to long term data sets of geometrically and radiometrically corrected satellite images. Radke \textit{et al}. \cite{radke2005image} presents a comprehensive overview of this class of change detection methods. More recently, Deep CNN based models have been used to classify the difference images for change detection in \cite{zhao2014deep}. Using Siamese networks for pixel wise change detetion has been studied in \cite{daudt2018fully}. Daudt \textit{et al}. \cite{Daudt_2019_CVPR_Workshops} use an iterative refinement and training procedure to learn a change detection model from noisy data. Recurrent Neural Networks have also been used for Change Detection in Multi-Temporal data \cite{papadomanolaki2019detecting}.

\subsection{Building Damage Assessment}

While the literature on building footprint segmentation and change detection is very extensive, the problem of building damage assessment from satellite imagery has only recieved limited attention. Xu \textit{et al}. \cite{xu2019building} investigate a two stage architecture for detecting damaged buildings. A Faster R-CNN \cite{ren2015faster} model is trained to detect building tiles, and a change detection network is trained as a binary classifier on pre and post disaster building tile pairs. Various simple CNN architectures are tried for the change detection network, and the architecture using feature differences is found to be the best. Gupta \textit{et al}. \cite{gupta2019xbd} provide baseline results for the xBD dataset. They adopt a U-Net based model originally designed to detect building footprints in SpaceNet for their first stage. Additionally, for change detection, they use a two-stream classification approach. 

A crucial limitation of these methods is the use of separate stages for detecting buildings and damage classification. This necessitates the use of greedy stage wise training, which   makes it impossible for the model to benefit from multi-task supervision. RescueNet explores the possibilities offered by a joint segementation and damage assessment framework and aims to close this gap in the literature. 


\section{RescueNet Model} \label{model}

\begin{figure*}[!h]
    \advance\leftskip-1cm

    \includegraphics[]{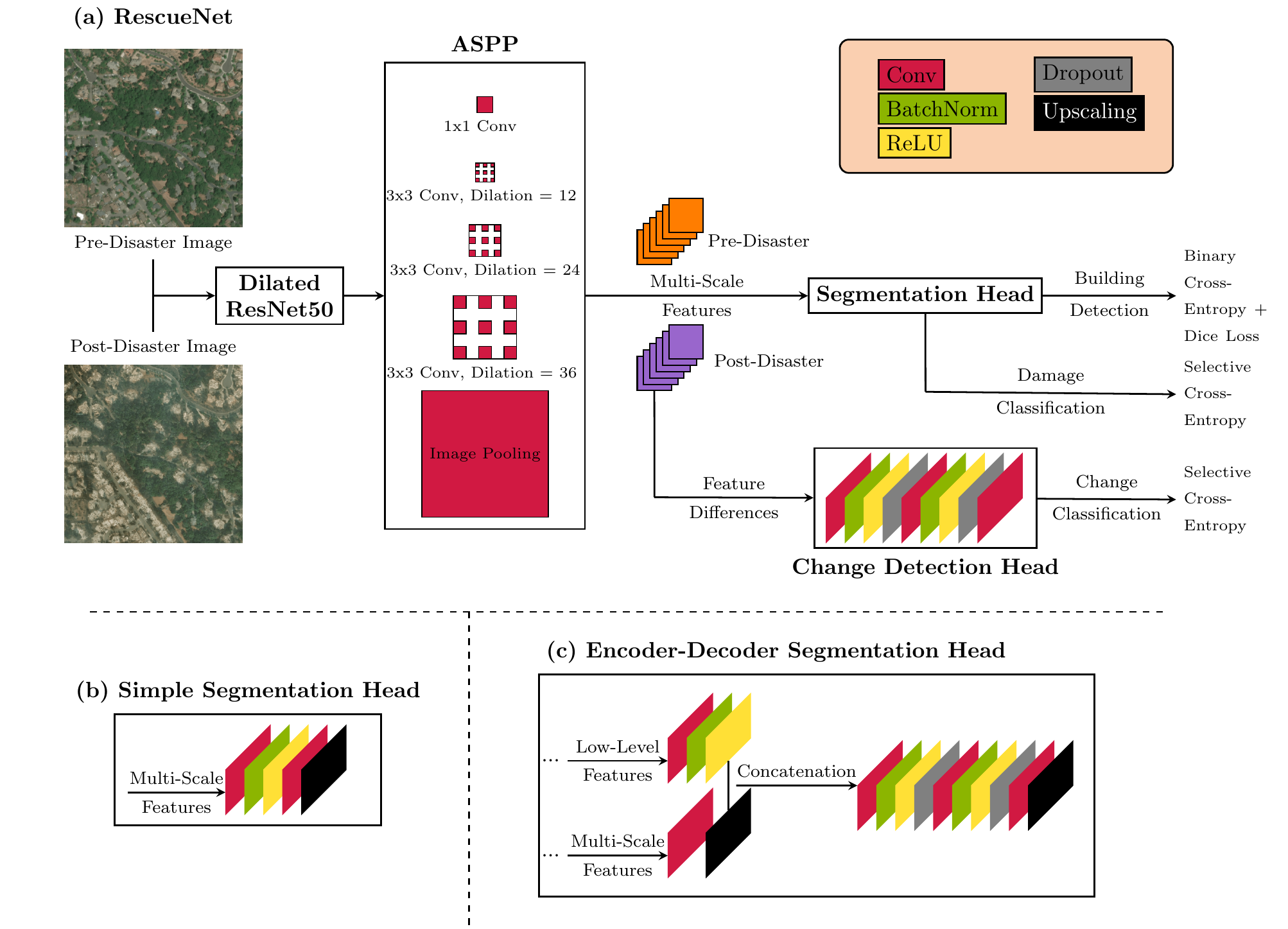}

    \caption{\textbf{(a) RescueNet Model.} The backbone of the network is a Dilated ResNet, which produces features at 1/8th input image resolution. \`{A}trous Spatial Pyramid Pooling module uses dilated convolutions of strides 12, 24 and 36, along with image pooling to generate multi-scale features. Segmentation head can have either the simple convolutional design or follow the encoder-decoder style. The segmentation head makes predictions independently for pre- and post-disaster images. Change Detection head applies Convolutional layers with batch normalization and ReLU activations followed by an output 1x1 Convolution to the difference of post- and pre-disaster features. The xBD dataset has 4 different damage classes (C=4), while building segmentation is a binary problem. \textbf{(b) the simple segmentation head.} \textbf{(c) the encoder-decoder style segmentation head.} }
    \label{fig:model}
\end{figure*}


RescueNet is an unified model for building segmentation and damage classification. A schematic blockgiagram showing components of RescueNet's architecture grouped into logical modules is provided in Figure \ref{fig:model}. The design of each of these modules and explanations for the design choices made therein are provided further on in this section. The design of the feature extractor and segmentation head draw on the semantic segmentation literature, specifically DeepLabv3 \cite{chen2017rethinking} and DeepLabv3+ \cite{chen2018encoder}, adapting them to the task of segmenting buildings, and augmenting it with the capability of change detection across multi-temporal images.

\subsection{Extracting Image Features}

The goal of the first stage of the network is to extract multi-scale image features at a high spatial resolution. In order to preserve higher spatial resolution in the features RescueNet adopts a backbone architecture consisting of a ResNet50 \cite{he2016deep}, with dilated convolutions replacing strided convolutions beginning from the second ResNet Blocks \cite{yu2017dilated}. This results in features having 1/8th the spatial resolution of the input image along both height and width. To obtain multi-scale features \'{A}trous Spatial Pyramid Pooling (ASPP) module is used on top of the backbone CNN features. The ASPP module uses image pooling and dilated convolutions of kernel size 3 and dilation sizes of 12, 24 and 36 to obtain features at 4 different scales. 

\subsection{Segmentation Head}

 We experiment with two possible architectures for the segmentation head. The simple architecture consists of a convolution block, whereas the Encoder-Decoder architecture additionally has skip connections from the backbone network's lower level features. In both cases these layers are followed by upsampling to make segmentation predictions at full image resolution. The Segmentation head classifies the pixels of the pre- and post-disaster images independently, hence for damage classification, it  only processes post-disaster imagery.

\subsection{Change Detection Head}

In order to utilize the temporal (post and pre-damge) features available from the input, we design an additional change detection head. The change detection head is a simple Convolution block with 3 layers, and utilizes as its input the difference of multi-scale features from the pre- and post disaster images. We utilize the difference of features as the amount of change is expected to be associated with the distance in the semantic feature space. The change detection head classifies each pixel into the four damage categories, and the predictions made for background pixels are ignored while training and during inference. 



\subsection{Loss Functions}

Choosing a good loss function is very important and hence we experiment with a variety of relevant loss functions.
To begin with we use the simple Cross-Entropy Loss on top of the segmentation output to train the network. However, the cross-entropy loss is not well suited for this problem because the network needs to classify pixels at 2 different levels: building localization and damage classification. With this insight in mind, we developed a localization aware loss function, which accounts for the hierarchical nature of the problem. The localization aware loss consists of a Binary Cross Entropy loss for building segmentation, and a foreground only selective Categorical Cross-Entropy loss for damage classification. Since this loss matches the structure of the problem better, we expect it to perform significantly better than plain Cross-Entropy. Formally, the localization aware loss function is defined in Equation \ref{eq:locawareloss}. In the equation,  $y_{il}$ is the ground-truth binary localization label indicating whether the current pixel belongs to a building or not, $y_{ik}$ is the damage classification label for class $k$, and $\hat{y_{il}}$ and $\hat{y_{ik}}$ are the predicted values for them. The damage classification loss is summed over the set of all damage classes $C$.


\begin{equation}
 \label{eq:locawareloss}
  L(y_i, \hat{y_i}) =
  \begin{cases}
    -log(\hat{y_{il}}) + \sum_{k\in C} -log(\hat{y_{ik}}) & \text{if $y_{il}=1$} \\
    -log(1 - \hat{y_{il}}) & \text{if $y_{il}=0$}
  \end{cases}
\end{equation}


Additionally, in order to predict crisper building boundaries we also employ with using the Dice Loss \cite{milletari2016v} \cite{deng2018learning} for building segmentation. The Dice Loss is based on the Sørensen–Dice coefficient: 

\begin{equation}
Dice Loss(y_{il}, \hat{y_{il}}) =  \frac{2 |y_{il}\cap\hat{y_{il}}|}{|y_{il}| + |\hat{y_{il}}|}.
\end{equation}


Since $y_{il}$, the ground truth building mask label is binary, the intersection term in the numerator can be computed as the product of ground-truth and prediction, and the denominator is the sum of the absolute values, which means the loss is piece-wise differentiable and hence trainable.



\section{Dataset} \label{dataset}

We use xBD \cite{gupta2019xbd}, which is the only large-scale public dataset for building segmentation and damage assessment to benchmark our method. xBD has been collected for the purpose of aiding research that results in technology to assist Humanitarian Aid and Disaster Relief Efforts. It consists of high quality building segmentation and damage assessment annotations for high-resolution satellite imagery, collected before and after 19 disaster events (such as floods, volcanic eruptions, earthquakes, and hurricanes) spread across the world. The dataset uses polygons to reprsent building segments and a novel 4 point damage scale. The dataset consists of geo-registered pairs of pre- and post-disaster images of size 1024 pixels x 1024 pixels with building polygons and 4-class damage labels provided for each building. We use the train split of xBD to train and test our method as the test set annotations are not publicly available yet. The training data is divided by the original authors into Tier1 and Tier3 data, with each tier correspond to a different set of disaster events. We split off about 10\% of the Tier1 data into a validation set using stratified sampling across disaster events to ensure a representative sample. Our models are trained on the Tier 1 and Tier3 Train set and tested on the Tier1 validation set. Sizes of each dataset split can be found in Table \ref{table:dataset}. For reproducability, the list of data samples in the validation split will be released with the supplementary material. 

\begin{table}[h!]
\caption{Size of xBD splits}
\centering
\begin{tabular}{ |c|c| } 
\hline
 Data Split & Image Pairs \\ 
 \hline
 Tier1 Train & 2,495 \\ 
 Tier3 Train & 6,369 \\ 
 Tier1 Validation & 304 \\
 \hline
\end{tabular}
\label{table:dataset}
\end{table}

\section{Results} \label{results}

In this section we present the results of benchmarking our model on the xBD dataset. We  first present  results from our model for some samples from the validation set, then define the XView Challenge metric which is used as the primary quantitative metric, then present ablation studies in order to study the impact of various components of our method, followed by a comparison of our results with the baseline.

\subsection{Qualitative Results}

\begin{figure*}[!h]
\centering

\subfloat{\includegraphics[width=1.4in]{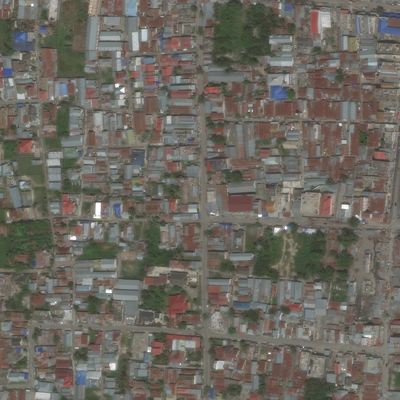}}
\hfill
\subfloat{\includegraphics[width=1.4in]{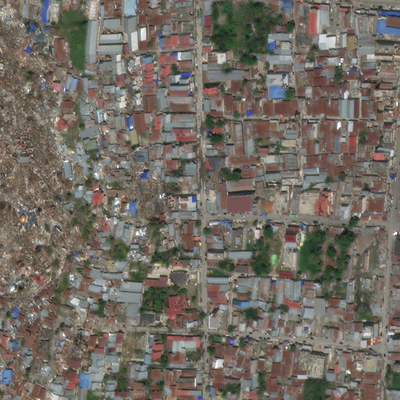}}
\hfill
\subfloat{\includegraphics[width=1.4in]{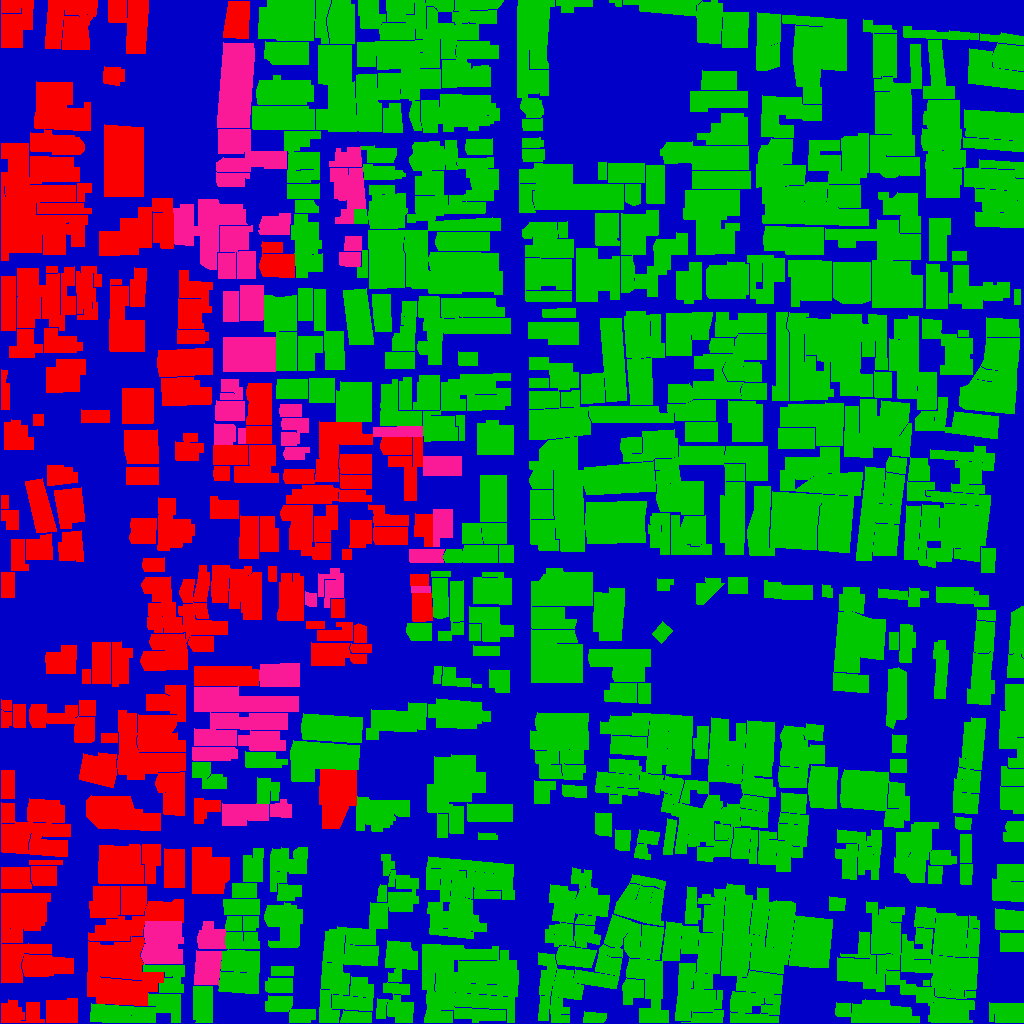}}
\hfill
\subfloat{\includegraphics[width=1.4in]{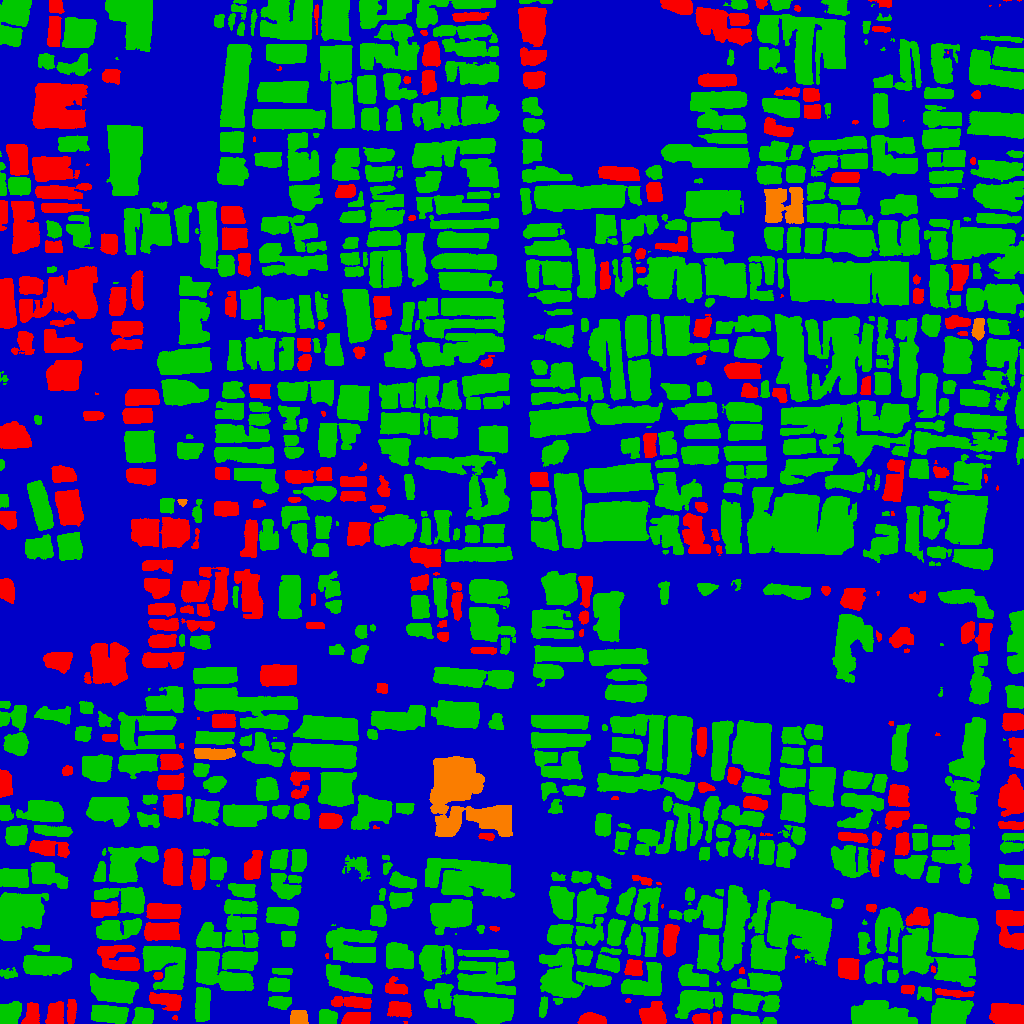}}
\hfill
\subfloat{\includegraphics[width=1.4in]{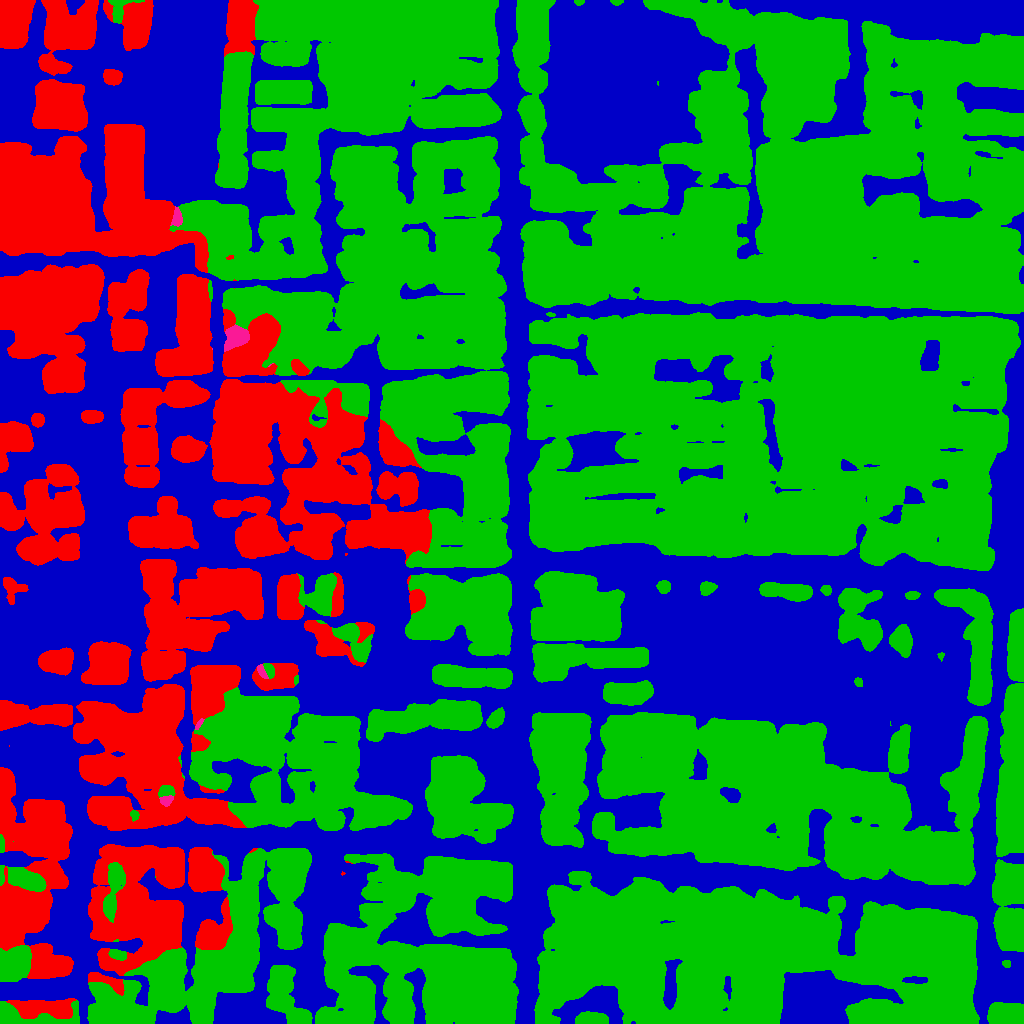}} \\[1ex]

\subfloat{\includegraphics[width=1.4in]{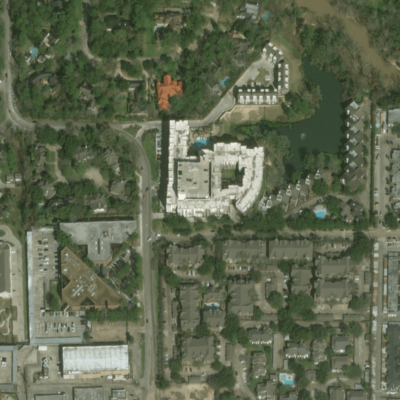}}
\hfill
\subfloat{\includegraphics[width=1.4in]{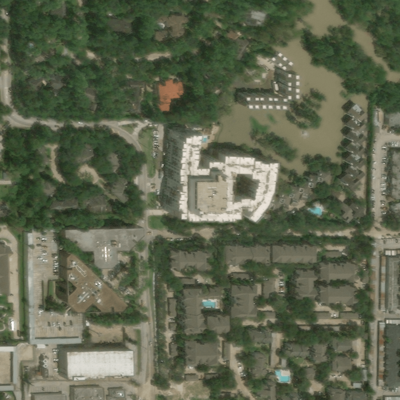}}
\hfill
\subfloat{\includegraphics[width=1.4in]{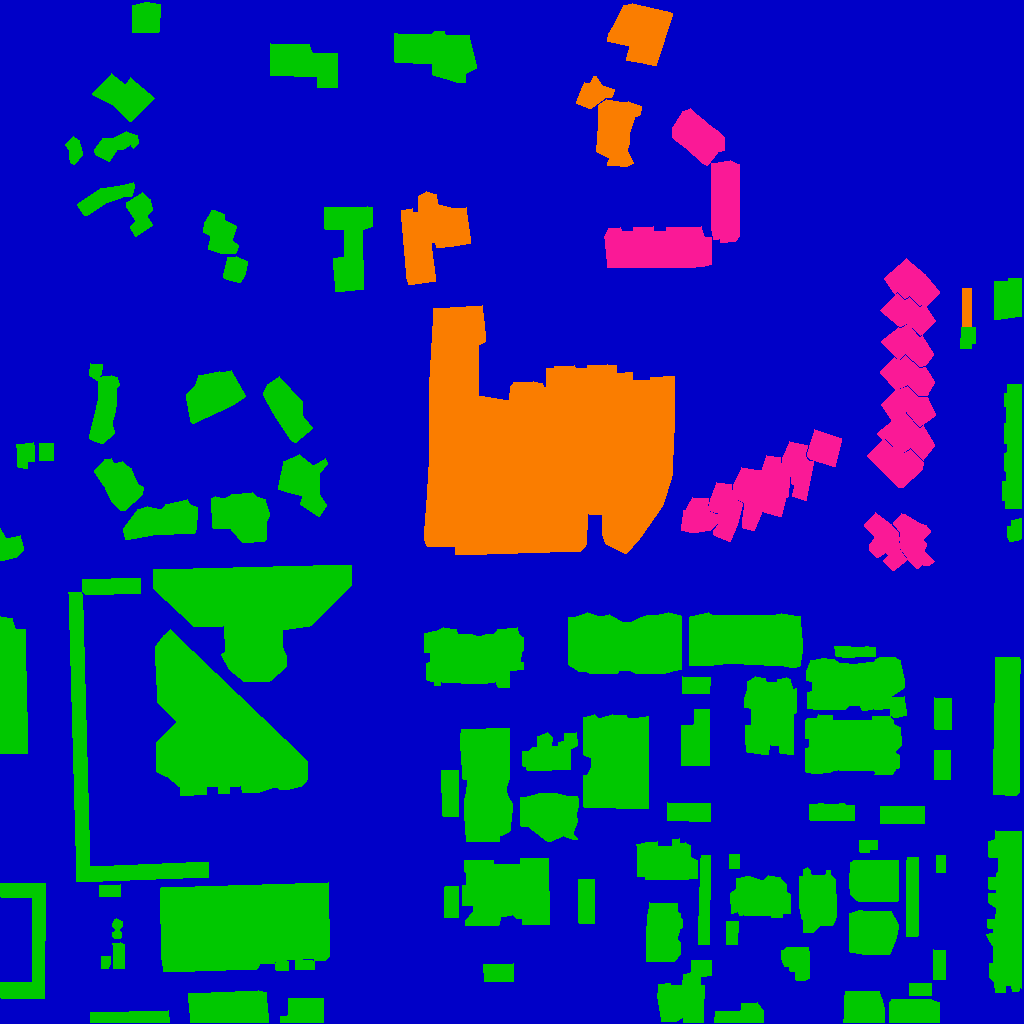}}
\hfill
\subfloat{\includegraphics[width=1.4in]{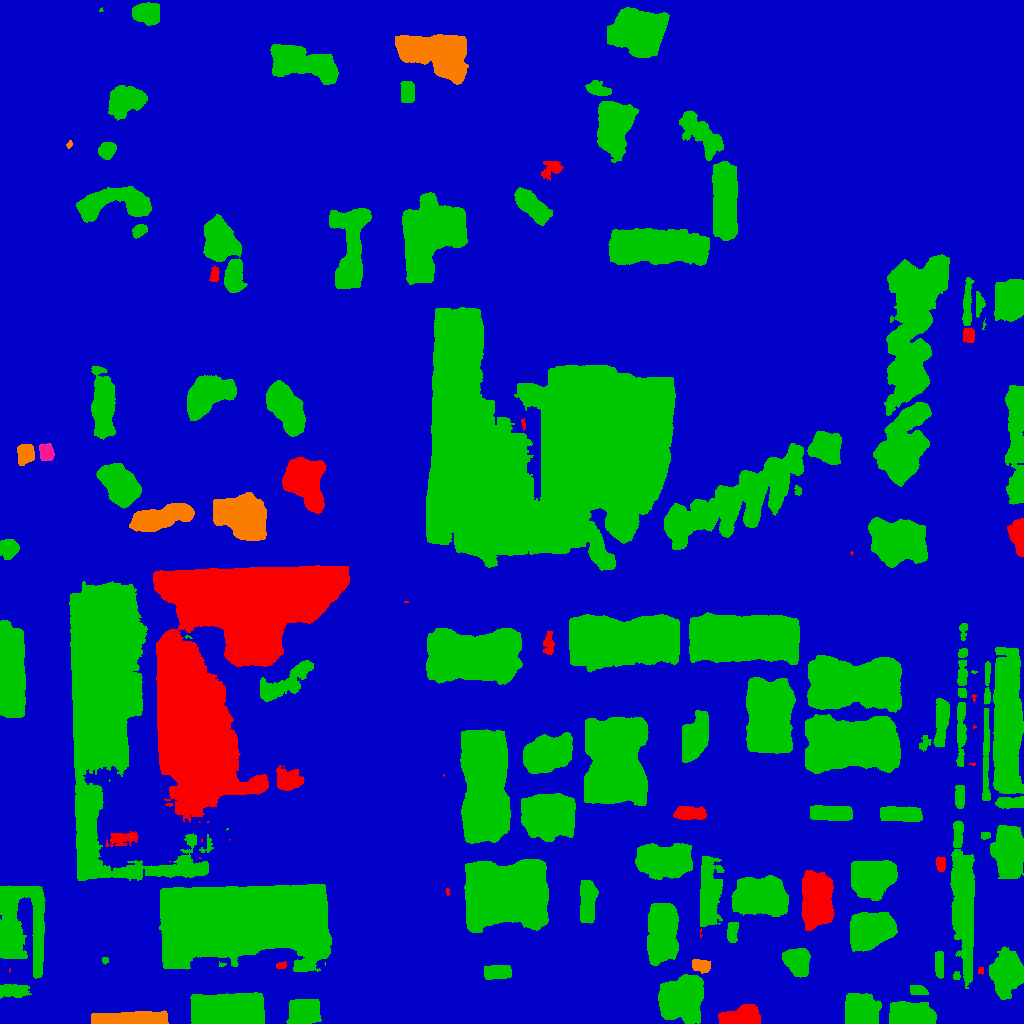}}
\hfill
\subfloat{\includegraphics[width=1.4in]{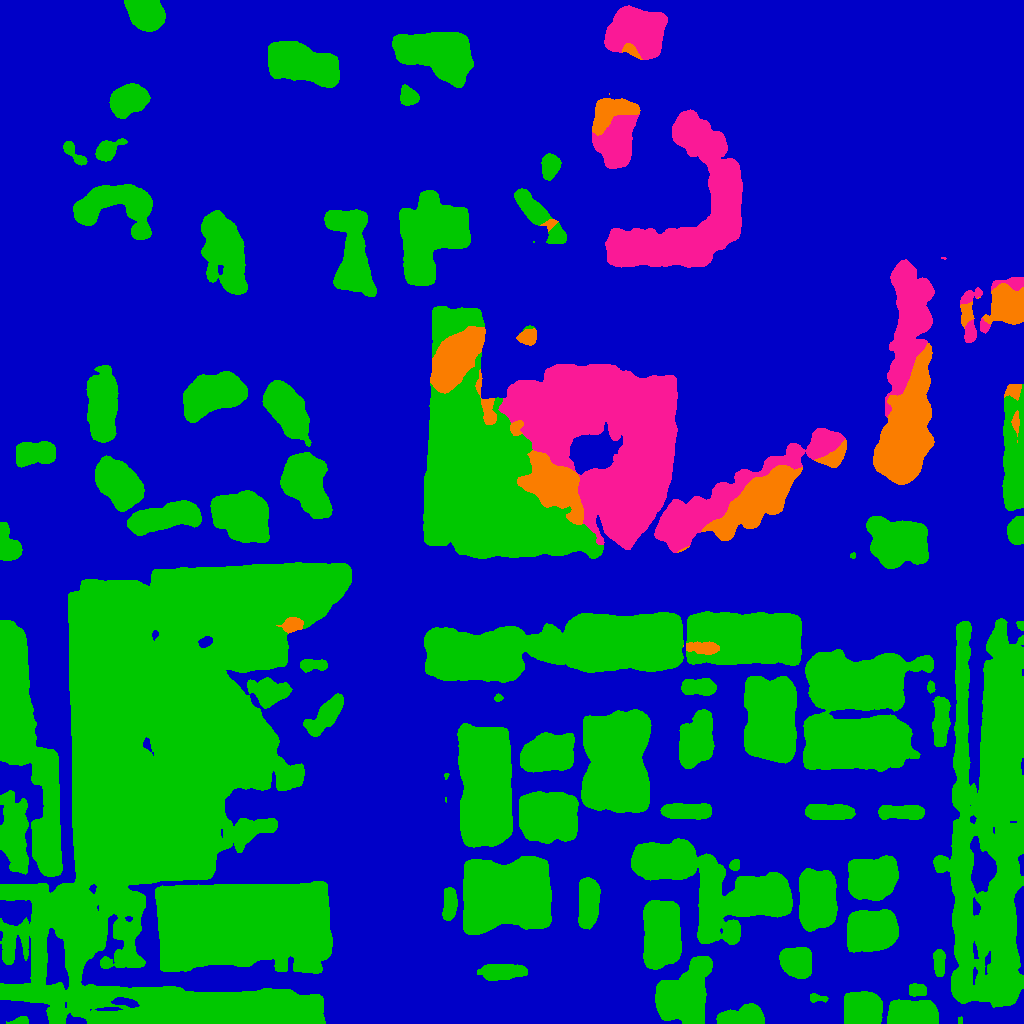}} \\[1ex]

\subfloat{\includegraphics[width=1.4in]{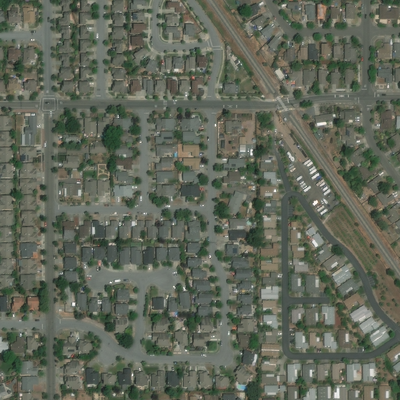}}
\hfill
\subfloat{\includegraphics[width=1.4in]{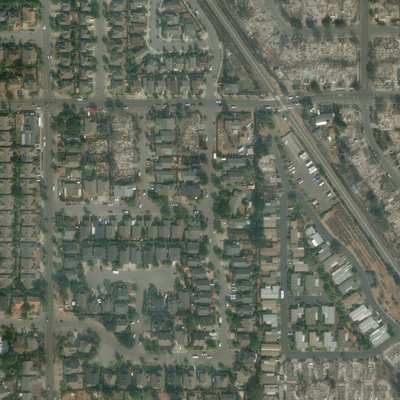}}
\hfill
\subfloat{\includegraphics[width=1.4in]{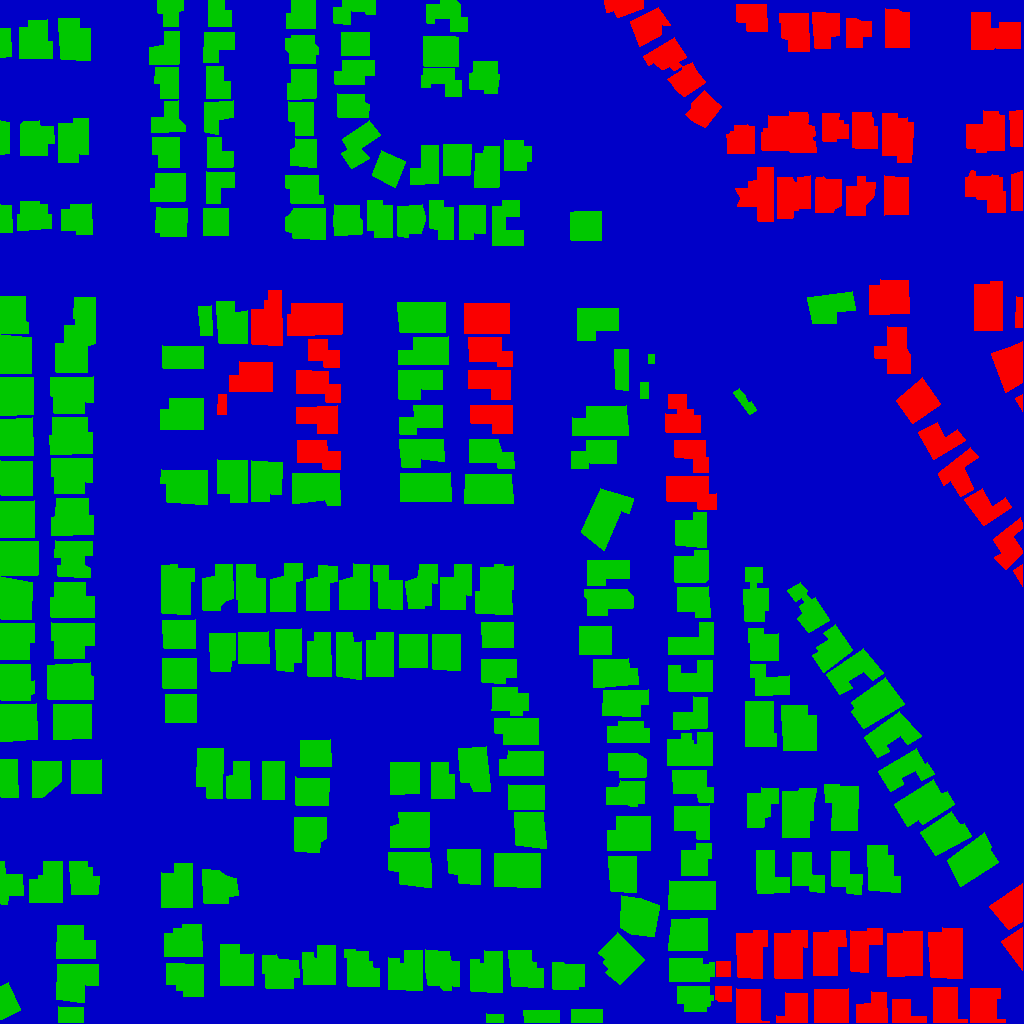}}
\hfill
\subfloat{\includegraphics[width=1.4in]{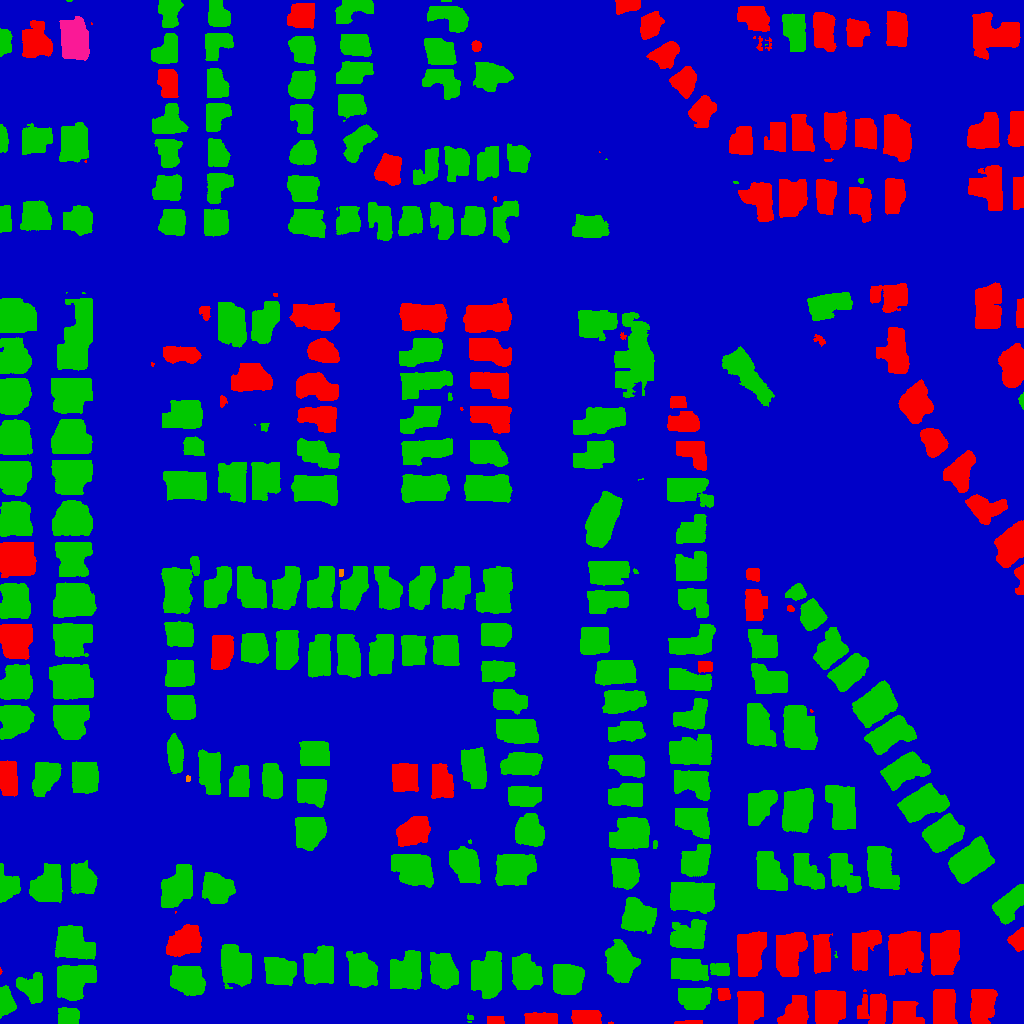}}
\hfill
\subfloat{\includegraphics[width=1.4in]{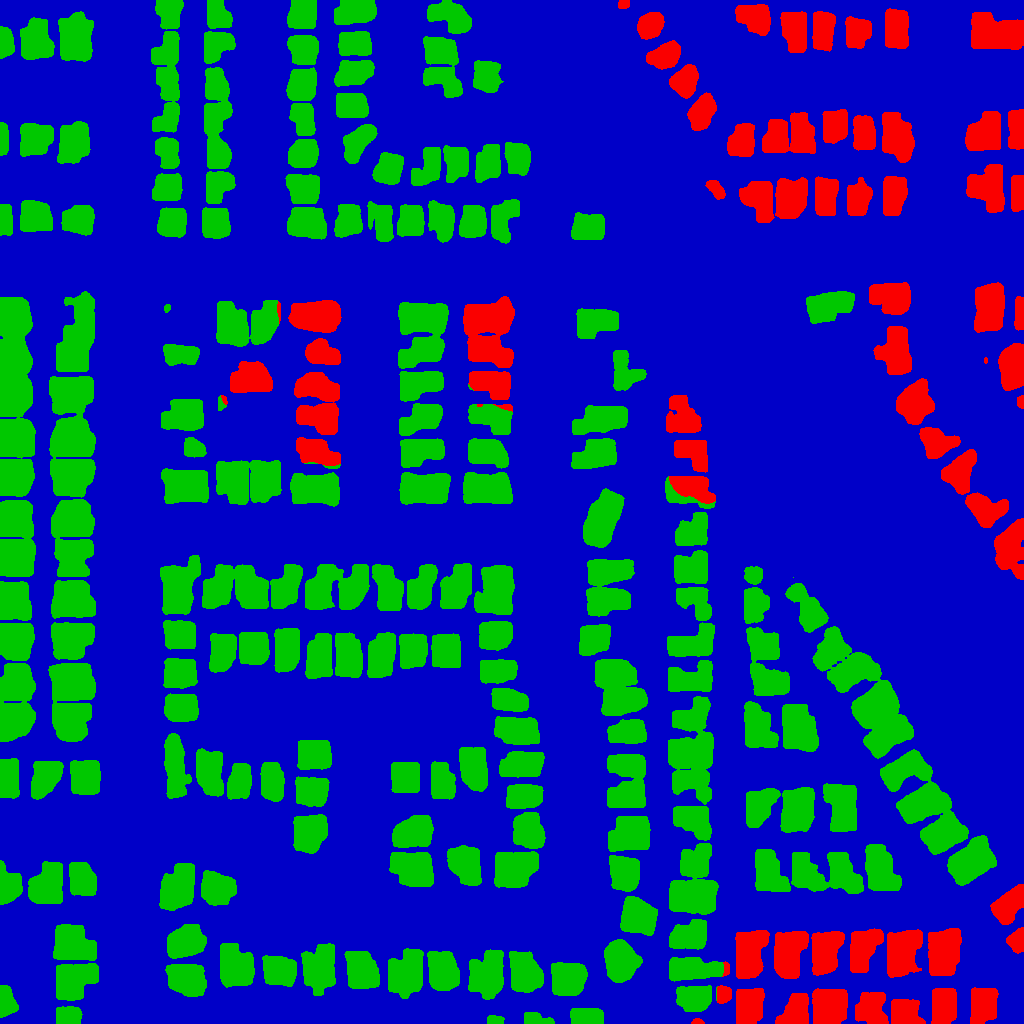}} \\[1ex]

\subfloat{\includegraphics[width=1.4in]{figures/intro_pre_image.png}}
\hfill
\subfloat{\includegraphics[width=1.4in]{figures/intro_post_image.png}}
\hfill
\subfloat{\includegraphics[width=1.4in]{figures/intro_post_gt.png}}
\hfill
\subfloat{\includegraphics[width=1.4in]{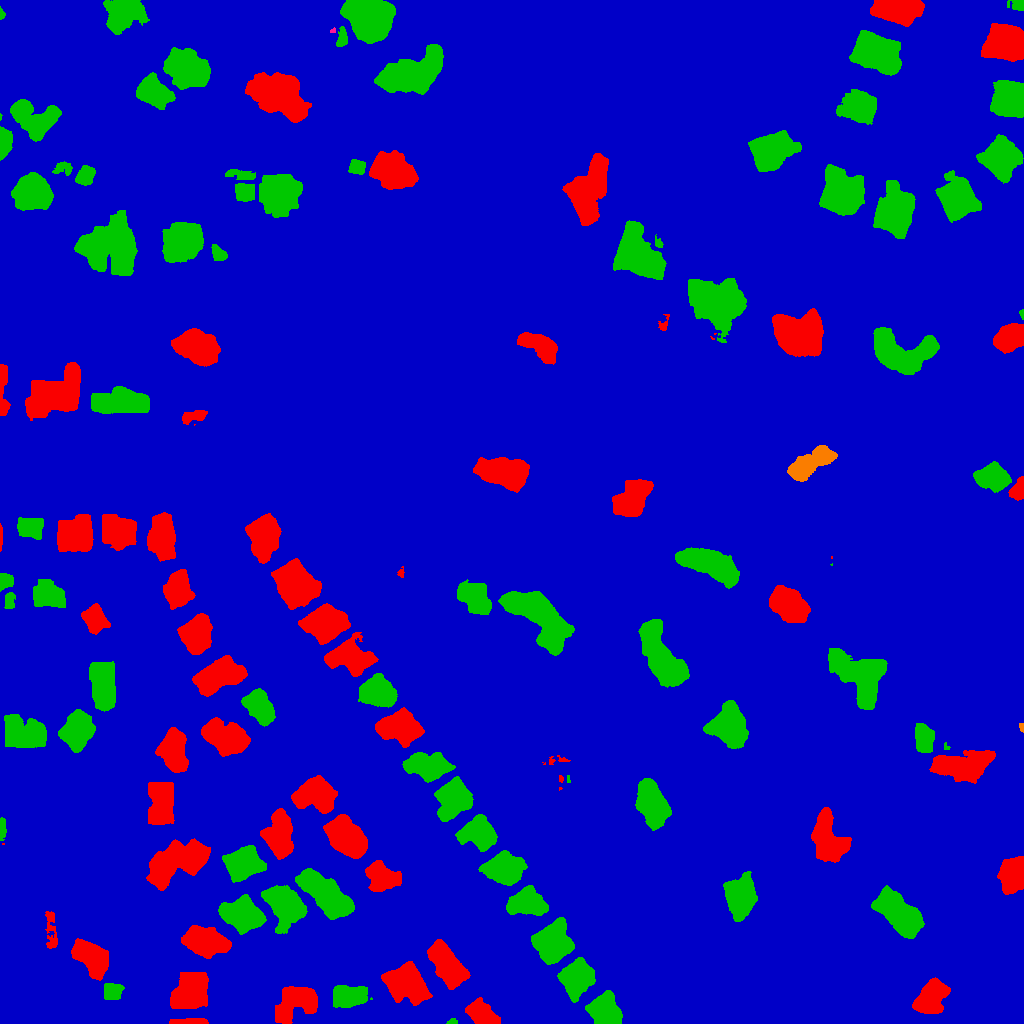}}
\hfill
\subfloat{\includegraphics[width=1.4in]{figures/intro_combined_pred_epoch_58.png}} \\[1ex]

\setcounter{subfigure}{0}

\subfloat[Pre-Disaster]{\includegraphics[width=1.4in]{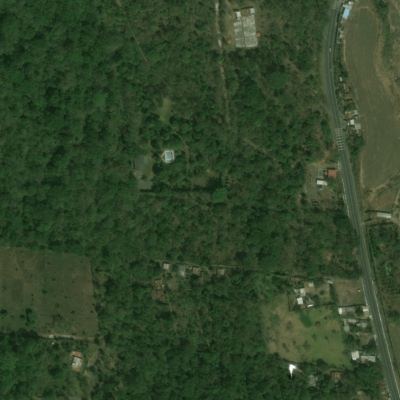}}
\hfill
\subfloat[Post-Disaster]{\includegraphics[width=1.4in]{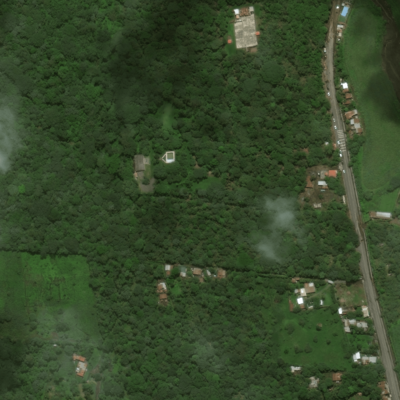}}
\hfill
\subfloat[Ground Truth]{\includegraphics[width=1.4in]{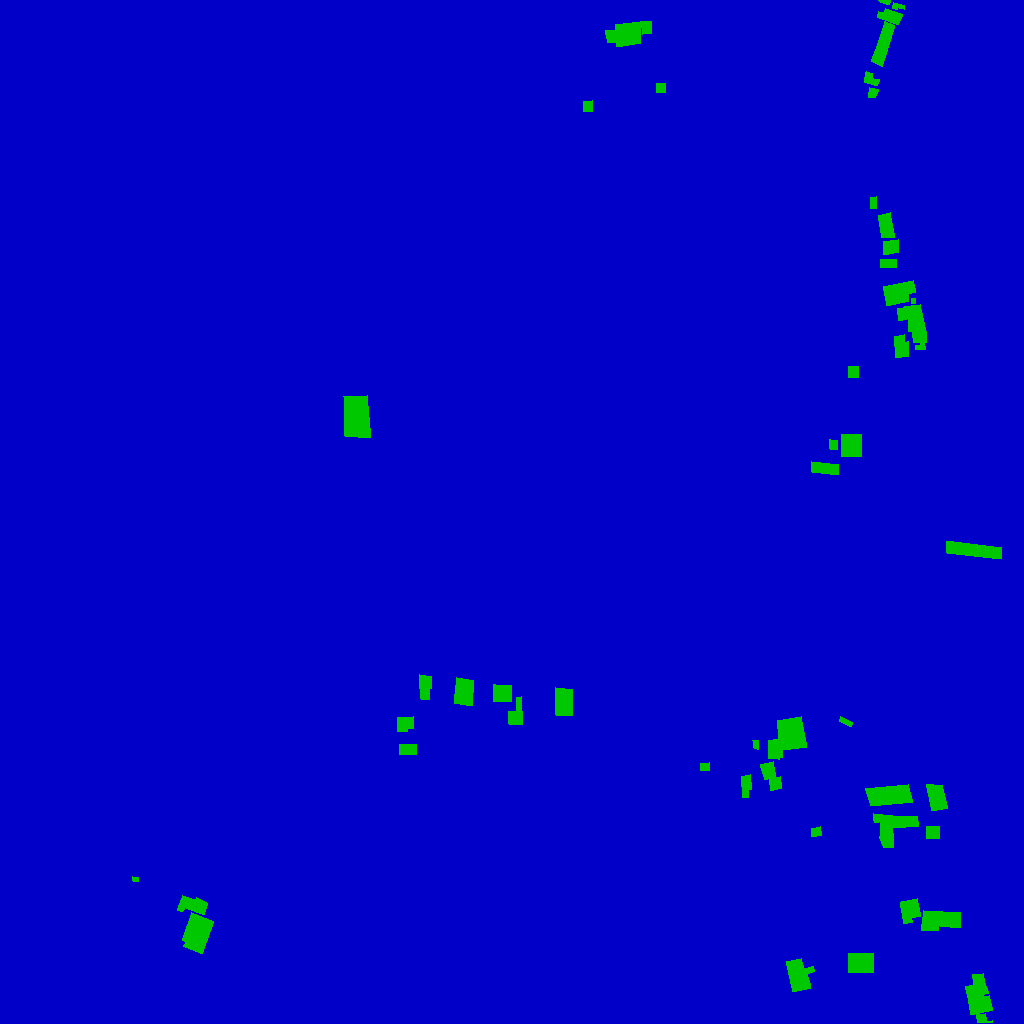}}
\hfill
\subfloat[Baseline Prediction]{\includegraphics[width=1.4in]{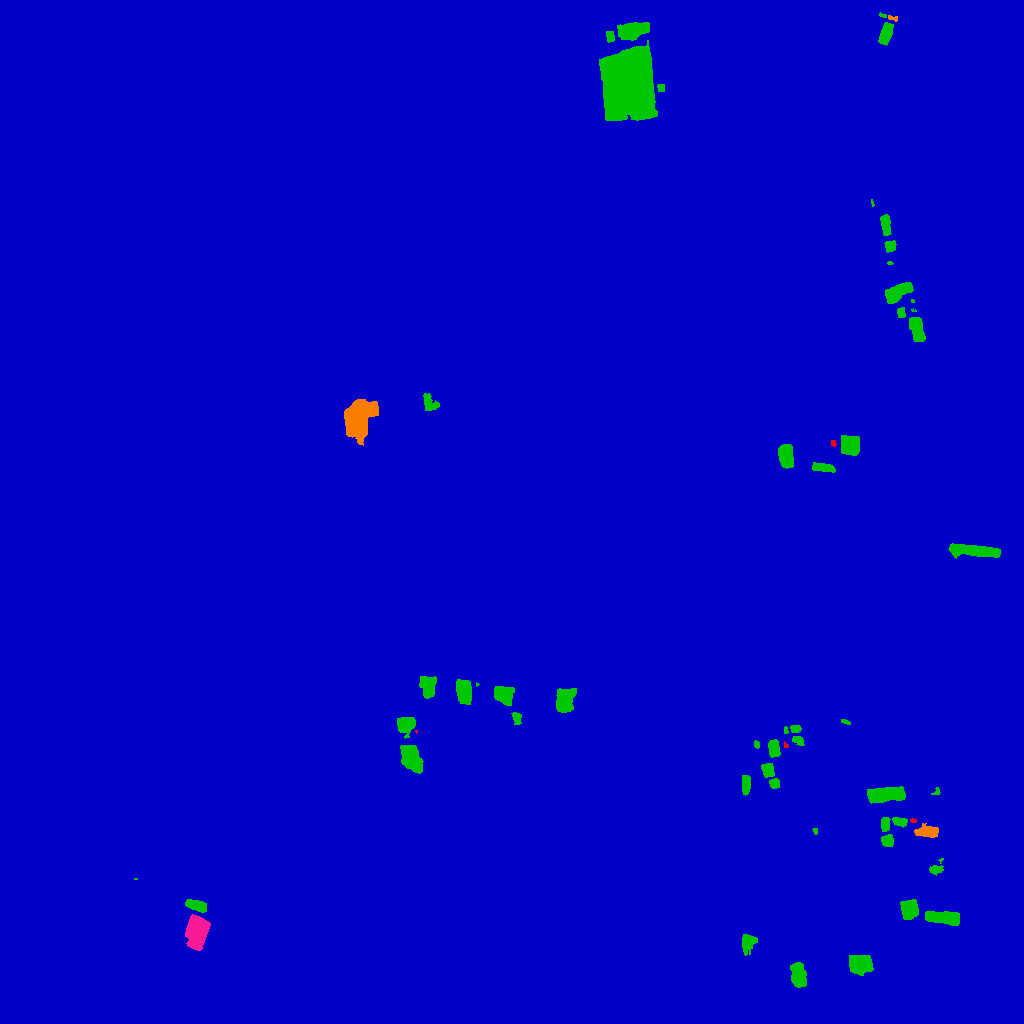}}
\hfill
\subfloat[Prediction (Ours)]{\includegraphics[width=1.4in]{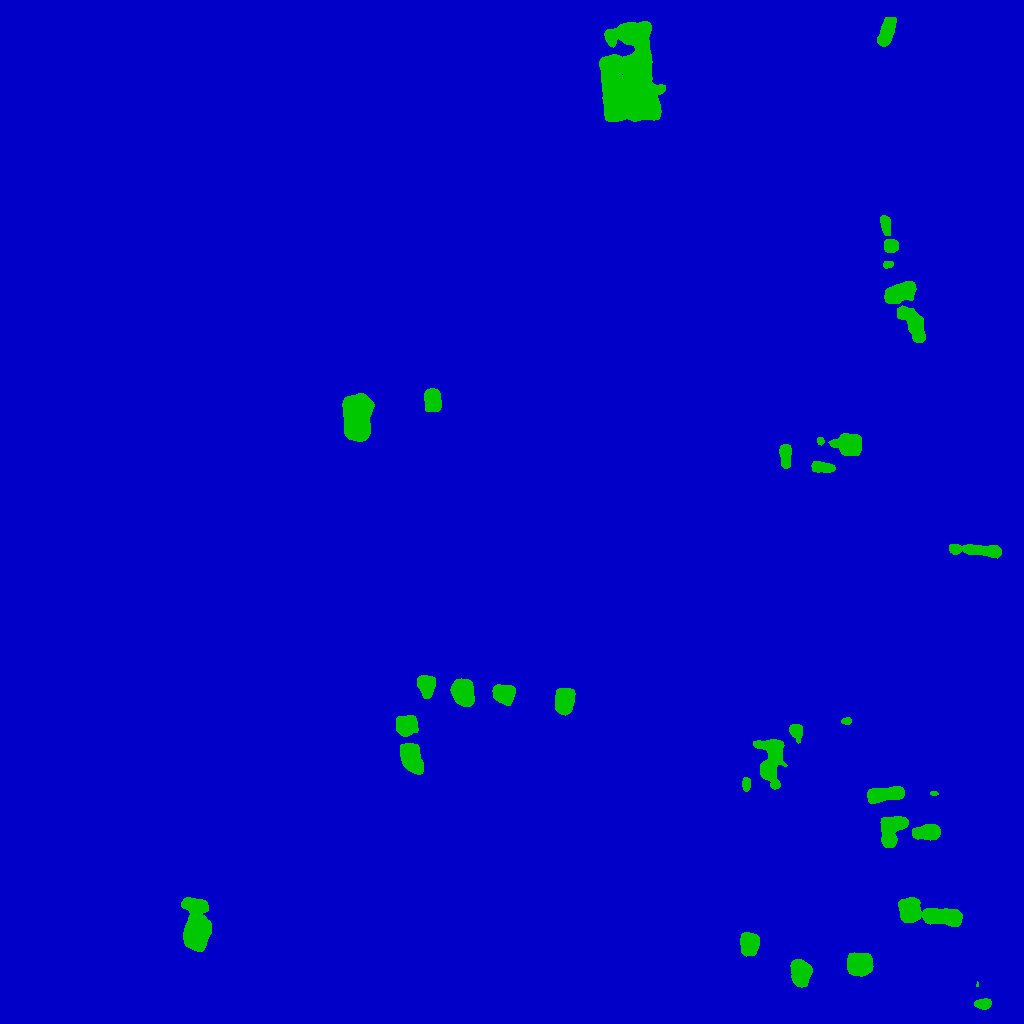}} \\[1ex]


\caption{Qualitative Results for a diverse set of examples from the Validation set. \textbf{From Left to Right in each row:} Pre-Disaster Image, Post-Disaster Image, Ground Truth Labels, Predicted labels (Baseline Model), Predicted labels (RescueNet). \textbf{From Top to Bottom:} A very high density urban area severely hit by a Tsunami, Partially flooded high density region with diverse building sizes, suburban region with near uniform building sizes and moderate density with localized housing damage due to forest fires, similar suburban region with widespread damage due to forest fires, and finally a rural area with small and sparse buildings. The baseline network has a large of number of false positives and false negatives in the examples with significant numbers of buildings. \textbf{Segmentation mask colors:} Blue (areas outside buildings), green (undamaged buildings), red(completely destroyed buildings), pink(buildings with major damage) orange(buildings with minor damage). Best viewed in high resolution color.}


\label{fig:qual}
\end{figure*}


In Figure \ref{fig:qual} we present some qualitative results on a small but diverse sample of the dataset. The selected images are from areas with significant diversity in density of buildings and type of damage. As can be noted from these results, RescueNet appears to be significantly better than the baseline model at predicting the damage levels. The baseline model makes quite a few false positive and false negative errors that are eliminated by our model. Significant variation can be observed in the pattern of damage caused by different kinds of disasters, with flooding affecting neighbouring buildings to widely varying extents, whereas disasters like Tsunamis which strike with concentrated force destroy a contiguous region. The qualitative results also suggest two sources for the remaining error of our model: first, extremely large buildings not being recognized as one unit by the damage classification head, and secondly, the building segmentation head not being able to separately resolve neighbouring buildings that are extremely close to each other. 


\afterpage{\clearpage}

\subsection{Metrics}

We report results using the XView2 Challenge metric\footnote{\url{https://xview2.org/challenge}}, a weighted average of the building segmentation F1 score and the harmonic mean of class wise damage classification F1 scores:


\begin{equation}
Score = 0.3 * F1_{loc} + 0.7 * \frac{n}{\frac1{F1_{cls1}} + \cdots + \frac1{F1_{clsn}}}
\end{equation}

Here, $F1_{loc}$ is the F1-Score for building segmentation and $F_{cls1} \cdots F_{clsn}$ are the damage classification F1-scores for each damage class. This metric is very challenging because it heavily penalizes overfitting to overepresented classes and the xBD dataset is heavily skewed. 


\subsection{Ablation Studies}

In order to study the impact of our design choices, we carry out ablation studies with different loss functions, segmentation heads and with or without the change detection head. The result for each of these experiments are discussed in this section. 
\subsubsection{Losses} The impact of using different loss functions can be observed from Table \ref{table:ablationstudies}. Using the localization aware loss provides a substantial improvement over the standard cross-entropy loss and adding in the Dice Loss provides a further, smaller boost.

\begin{table}[h!]
\caption{Ablation Studies} 

\centering
\begin{tabular}{ |c|c| } 
\hline
 {\textbf{Loss Function}} & {\bf Score}\\ 
 \hline
 Cross-Entropy Loss &  0.69 \\ 
 Localization Aware Loss & 0.75 \\ 
 \textbf{Localization Aware Loss + Dice Loss} & \textbf{0.77} \\
 \hline
  \multicolumn{2}{|c|}{\textbf{Segmentation Head Architecture}} \\ 
 \hline
 Simple (Convolution+Upscaling) &  0.74 \\ 
 \textbf{Encoder-Decoder} &  \textbf{0.77} \\ 
 \hline
  \multicolumn{2}{|c|}{\textbf{Change Detection Head}} \\
 \hline
 Without change detection head &  0.76 \\ 
 \textbf{With change detection head} & \textbf{0.77} \\ 
 \hline

\end{tabular}
\label{table:ablationstudies}
\end{table}

\subsubsection{Segmentation Head Architecture} 

Using the Encoder-Decoder segmentation head provides a boost in performance over the simple segmentation head, as can be seen in Table \ref{table:ablationstudies}.  


\subsubsection{Change Detection Head}

Finally, we can separate out the impact of using the change detection head which provides a boost of about  1 percentage point. See Table \ref{table:ablationstudies}.


\subsection{Generalization} 

In order to verify the model's ability to generalize across different geographical regions and disaster types, we carry out a couple of additional experiments where the model is trained on Tier 1 training data and tested on Tier3 data, which belongs to a different set of disasters, and hence is a good measure of generalization ability. The results for these experiments can be seen in Table \ref{table:generalization}. The results indicate that the model is able to generalize on the building segmentation task, but suffers significant degradation on the damage assessment task.


\begin{table}[h!]
\caption{Generalization across regions and disaster types}
\centering
\begin{tabular}{ |c|c|c|c|c| } 
\hline
& & \multicolumn{3}{|c|}{\textbf{Score}} \\ 
\hline
 {\bf Train Set} & {\bf Test Set} & {\bf Localization} & {\bf Damage} & {\bf Overall} \\ 
 \hline
 Tier1 Train & Tier1 Valid. & 0.79 & 0.60 & 0.66 \\ 
 Tier1 Train & Tier3 & 0.77 & 0.37 & 0.50 \\ 
 \hline
\end{tabular}
\label{table:generalization}
\end{table}
\subsection{Comparison}

Baseline Results for xBD dataset are reported in \cite{gupta2019xbd}. As previously noted, this is a two stage approach, where the first stage is a U-Net based building segmentation model and the second stage is two-stream damage classification model that operates on building tiles. Results reported in the baseline Paper are for the Tier 1 test set, which is not available to us at present, hence for a fair comparison we train the baseline with our dataset split using code provided by the authors and using the same hyper parameters.

Table \ref{table:baselinecompare} provides a comparison of the overall score achieved by our model against the score achieved by the baseline model. Results obtained by training the baseline building segmentation model and damage classification model are shown in Table \ref{table:baselinestats} along with results reported in the original paper, as can be seen, we are able to closely reproduce the baseline results. Classwise metrics for the baseline damage classification model are compared with our results in Table \ref{table:baseline}. RescueNet is able to correctly classify the intermediate damage levels an order of magnitude better than the baseline.

\begin{table}[h!]
\caption{Overall Results (XView2 Metric)}
\centering
\begin{tabular}{ |c|c|c|c| } 
\hline
 \textbf{Model} & \textbf{Localization Score} & \textbf{Damage Score} & \textbf{Overall Score}\\ 
 \hline
 Ours & 0.84 & 0.74 & \textbf{0.77} \\ 
 \hline
 Baseline \cite{gupta2019xbd} & 0.79 & 0.03 & 0.26 \\ 
 \hline
\end{tabular}
\label{table:baselinecompare}
\end{table}


\begin{table}[h!]
\caption{Results for component networks of baseline model}
\centering
\begin{tabular}{ |c|c|c| } 
\hline
 \textbf{Metric} & \textbf{Baseline (Our Split)} & \textbf{Baseline Reported in \cite{gupta2019xbd}} \\ 
 \hline
 \multicolumn{3}{|c|}{\textbf{Building segmentation network}} \\ 
 \hline
 Average IoU & 0.65 & 0.66 \\
 \hline
 \multicolumn{3}{|c|}{\textbf{Damage Classification network}} \\ 
 \hline
 Weighted F1 & 0.6274 & 0.2654 \\ 
 Macro-Averaged F1 & 0.3100 & 0.3204 \\ 
 Harmonic Mean F1  &  0.0761 & 0.0342 \\
 \hline
 
 \hline
\end{tabular}
\label{table:baselinestats}
\end{table}

\begin{table}[h!]
\caption{Class-wise Damage Classification F1-scores}
\centering
\begin{tabular}{ |c|c|c| }
\hline
    \textbf{Damage class} & \textbf{Baseline (Our Split)} & \textbf{RescueNet (Ours)}\\
    \hline
   undamaged     & 0.7211 & 0.8832 \\
       minor     & 0.0235 & 0.5628 \\
       major     & 0.0105 & 0.7711 \\
   destroyed     & 0.4262 & 0.8079  \\
   
   \hline
   Harmonic Mean & 0.0282 & 0.7348  \\
\hline
\end{tabular}
\label{table:baseline}
\end{table}

\section{Conclusion} \label{conclusion}
In this paper, we described a novel unified model for simultaneously segmenting buildings and assessing damage level caused to them by natural disasters using multi-temporal satellite imagery. We achieved this by designing a multi-headed architecture and hierarchical loss function suited to the composite task. We carried out ablation studies to quantify the impact of our loss functions, segmentation head architecture and change detection head, and showed that it was able to generalize across geographic regions and disaster types, while significantly outperforming the existing baseline. For future work on this problem, we plan to focus on better techniques for resolving separate building instances, which would help improve both building localization and damage assessment. 


\section*{Acknowledgment}

The authors would like to thank the organizers of the XView2 Challenge, which first brought this problem to our attention.




\IEEEtriggeratref{16}
\bibliographystyle{IEEEtran}
\bibliography{citations}
%



\end{document}